Research paper

# Development and Testing of a Wood Panels Bark Removal Equipment Based on Deep Learning

**Rijun Wang[a,b], Guanghao Zhang[a], Hongyang Chen[a], Xinye Yu[a], Yesheng Chen[a], Fulong Liang[a], Xiangwei Mou[a,*], Bo Wang[b,c,*]**

[a]*School of Teachers College for Vocational and Technical Education, Guangxi Normal University, Guilin 541004, China*
[b]*Key Laboratory of AI and Information Processing, Hechi University, Yizhou 546300, China*
[c]*School of Artificial Intelligence and Smart Manufacturing, Hechi University, Yizhou 546300, China*

ABSTRACT

Attempting to apply deep learning methods to wood panels bark removal equipment to enhance the quality and efficiency of bark removal is a significant and challenging endeavor. This study develops and tests a deep learning-based wood panels bark removal equipment. In accordance with the practical requirements of sawmills, a wood panels bark removal equipment equipped with a vision inspection system is designed. Based on a substantial collection of wood panel images obtained using the visual inspection system, the first general wood panels semantic segmentation dataset is constructed for training the BiSeNetV1 model employed in this study. Furthermore, the calculation methods and processes for the essential key data required in the bark removal process are presented in detail. Comparative experiments of the BiSeNetV1 model and tests of bark removal effectiveness are conducted in both laboratory and sawmill environments. The results of the comparative experiments indicate that the application of the BiSeNetV1 segmentation model is rational and feasible. The results of the bark removal effectiveness tests demonstrate a significant improvement in both the quality and efficiency of bark removal. The developed equipment fully meets the sawmill's requirements for precision and efficiency in bark removal processing.

## 1. Introduction

Wood panels, as a standardized material, offer a range of advantages including aesthetic appeal, environmental friendliness, and sustainability. Due to their excellent properties, wood panels are widely used in various fields such as construction, interior decoration, and craftsmanship [1-4]. However, obtaining wood panels with practical value typically requires a series of complex processing steps applied to raw logs. Wood cutting is a crucial step in the wood processing industry. Its primary purpose is to cut logs into wood panels suitable for production based on their intended use and requirements. Different cutting methods and processes result in wood panels of varying shapes, sizes, and qualities. Among the key considerations in wood cutting is the removal of bark from the panels. This step eliminates impurities, bark, and irregular surfaces, making the wood easier to process and handle, and providing a better foundation for subsequent processing steps. Furthermore, bark removal helps reduce the presence of non-cellular tissues in the wood, decreases its moisture content, and prevents decay and mold, thus minimizing resource waste. However, the quality and efficiency of bark removal significantly affect both the yield and quality of finished wood panels. Different bark removal methods also influence the utilization rate of the logs. In practical production, sawmills often pursue technological innovations in bark removal equipment to achieve high efficiency and low-cost methods, ultimately improving log utilization and producing more finished wood panels. Therefore, developing and testing bark removal equipment that meets the demands of engineering applications is of great practical significance.

In traditional bark removal operations, sawmills typically rely on simple equipment to perform this task. Such equipment usually consists of mechanical components, saw blades, motors, conveyor belts, and basic control buttons. The entire process is guided by the operator's experience in judging the appropriate cutting width for non-standard wood panels (typically wood panels with bark). Based on this judgment, the panels are placed into the trimming equipment's cutting channel, which corresponds to the desired width. The bark is then removed, and the panels are cut into standardized sizes (with bark removed and trimmed), ready for further processing into wood products.

During the bark removal process, the primary challenge for operators is to visually estimate and judge the accurate cutting width to avoid material waste and improve utilization. Additionally, because the cutting channels are fixed, operators must ensure that the placement of non-standard wood panels is parallel to the cutting channel. As a result, extensive technical training or practical experience is required for operators to perform this task effectively. One of the major issues with this method is that prolonged visual inspection can lead to distraction and eye strain[5]. In practical applications, the work efficiency is generally around 10-15 panels per minute, leading to a low bark removal efficiency. More importantly, the subjective nature of manual operations affects the outcome of bark removal. Due to the inherent subjectivity of manual judgment, there can be significant misalignment between the panel placement and the cutting position, as well as inaccurate sizing of the wood panels after bark removal. These issues result in several drawbacks. On one hand, they lead to substantial wood material waste; on the other hand, the finished wood panels may still retain bark, affecting the quality of the final product and potentially compromising subsequent wood processing. As a result, this manual bark removal method no longer meets the efficiency and quality standards of modern industrial production lines. The emergence of machine vision technology has helped address these shortcomings by significantly improving the precision and speed of inspection [6-8]. Consequently, to maximize production efficiency, the bark removal process is gradually shifting towards automation and intelligent systems [9].

* *Corresponding author.*
E-mail address:

With the advancement of machine vision inspection technology, it has gradually become an alternative to manual visual inspection in industrial production [10-12]. In the field of wood cutting, research on bark removal should focus primarily on how to apply image segmentation methods—such as threshold-based [13,14], region-based [15,16], edge-based [17], clustering-based [18,19], and classifier-based [20] methods—into bark removal equipment. The goal is to accurately identify and remove bark, thereby maximizing the utilization of raw materials. Compared to manual visual inspection, applying traditional machine vision techniques to bark removal equipment can, to some extent, accurately identify and segment non-standard wood panels with fixed bark edge positions, regular shapes, and a distinct color contrast with the wood panels. However, in actual factory production, producers still encounter issues such as incomplete or excessive bark removal, leading to poor quality consistency, subpar wood panel quality, and significant wood waste. These problems mainly arise because machine vision technology struggles to handle non-standard wood panels with irregular bark edge positions, irregular shapes, and bark colors similar to the wood panel itself.

In recent years, deep learning has achieved unprecedented advancements in image classification, object detection and recognition, and image segmentation technologies. Deep learning offers several distinct advantages. First, deep learning models can achieve high accuracy when handling image tasks, enabling more precise image classification, detection, and segmentation. Second, these models are highly adaptable and can process complex image data involving multiple objects, varying scales, and different orientations. Therefore, it has gradually been applied in the wood processing industry. In the field of wood defect identification, many frameworks based on different types of CNN models, such as CNN [21], SSD [22], Faster-CNN [23], Mask R-CNN [24], etc., have been derived for applications in wood surface defect recognition. In addition, deep learning has been applied to other aspects of the wood industry, e.g., wood classification [25], wood species identification [26], wood failure prediction [27], wood stiffness prediction [28], wood volume measurement [29], etc.. Bark removal, as a fundamental step in wood processing, remains an area where the application of deep learning methods is still in the exploratory stage, and practical results in this area are scarcely reported.

Building on the success of the aforementioned research, and addressing the technical challenges in bark removal, we aim to apply deep learning technology to the bark removal process. By integrating machine vision technology, we aim to tackle the precise identification and segmentation of bark. Our goal is to develop a deep learning-based bark removal system for use in industrial-scale production and to conduct validation experiments. To achieve this, we designed a wood panel vision acquisition system and integrated it into the existing bark removal equipment at the sawmill. Image data of wood panels with bark were collected in the actual industrial environment of the sawmill, labeled, and used to create a wood panel dataset for the training and evaluation of deep learning models. We applied vision detection algorithms to perform edge detection and fitting of the cuttable regions of the wood panels, including calculations of the cuttable width, panel pose, and the distance traveled relative to the cutting channel. Finally, the model trained offline was deployed onto the bark removal equipment, and experimental research on bark removal was conducted under actual sawmill conditions. The main contributions of this research are summarized as follows:

- A comprehensive semantic segmentation dataset for bark removal processing was established, facilitating both related research and practical applications in wood processing.

- A deep learning-based bark removal system was developed to meet the demands of factory-scale bark removal, improving the quality and efficiency of the process while reducing wood resource waste.

- Bark removal processing tests were conducted at the sawmill, and the results demonstrated that the developed bark removal equipment fully meets the requirements of real-world production.

The organization of this paper is as follows: Section 2 introduces the developed bark removal equipment based on deep learning and elucidates its working principle. Section 3 describes the design and functionality of the visual inspection system, along with the establishment of a semantic segmentation dataset for wood panels with bark. Section 4 details the process of utilizing BiSeNetV1 for the semantic segmentation of wood panels with bark, providing a thorough explanation of the key data calculations necessary for bark removal. Section 5 presents both comparative experiments conducted under laboratory conditions and testing experiments performed in a factory setting. The paper concludes in Section 6 with a summary of the findings and a discussion of future research directions and focal areas.

## 2. Wood Panels Bark Removal Equipment

The overall structure of the designed Wood Panels Bark Removal Equipment is shown in Fig. 1. It primarily consists of mechanical components such as the frame, feeding conveyor, vision inspection system, Wood Panels clamping and adjustment mechanism, cutting mechanism, discharge conveyor, and sorting mechanism for different specifications. The electrical control unit of the equipment is mainly composed of an industrial control computer, PLC, stepper motors, DC motors, an electrical control cabinet, and various sensors. The designed Wood Panels Bark Removal Equipment automates the entire bark removal process, including feeding, clamping, visual measurement, pose adjustment, multi-specification cutting, and sorting.

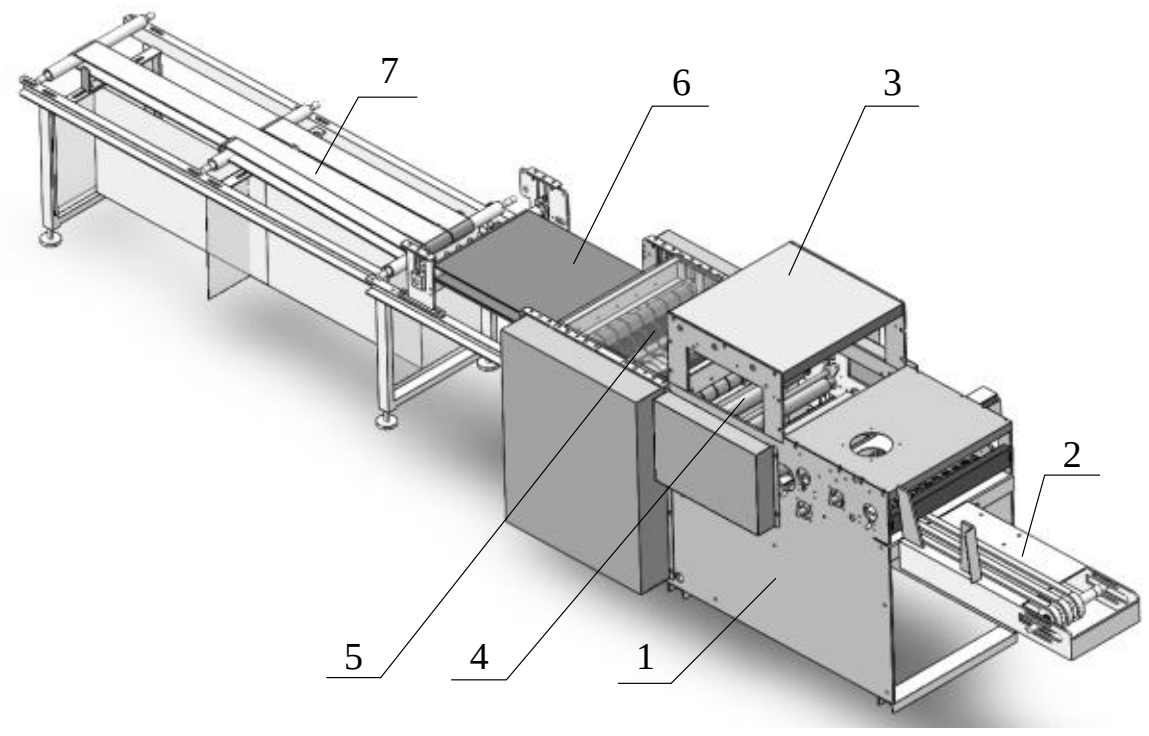

**Fig. 1. Schematic diagram of the wood panels bark removal equipment structure: 1 – Frame; 2 - Feeding conveyor; 3 - Vision inspection system; 4 - Wood panels clamping and adjustment mechanism; 5 - Cutting mechanism; 6 - Discharge conveyor; 7 - Sorting mechanism.**

Based on the actual bark removal requirements of sawmill operations, the working principle of this equipment can be summarized as follows: Operators (without the need for extensive technical training) stack the wood panels on the feeding conveyor mechanism. The feeding conveyor mechanism transports the wood panels one by one to the vision inspection position for processing and calculation, which primarily includes image acquisition of the wood panels, semantic segmentation, and the calculation of key data for bark removal. Based on the results of this processing and calculation, the clamping and adjustment mechanism adjusts the pose of the Wood Panels and transfers them to the corresponding cutting channel for bark removal. Among these steps, image acquisition, semantic segmentation, and the calculation of key data for bark removal are the core technologies of the wood panels bark removal equipment, directly determining the equipment's performance. The following sections will focus on addressing these key technological issues.

## 3. Construction of semantic segmentation dataset for wood panels with bark

### *3.1. Customized vision inspection system*

The vision inspection system is a critical component of the Wood Panel Bark Removal Equipment, serving as a key technological link for the rapid and accurate bark removal process. After thoroughly considering the production environment, requirements, and cost constraints of the sawmill, we determined the design and selection of the main hardware components. The framework of the vision inspection system is designed to maximize space, ensuring optimal imaging performance from the industrial camera mounted on it and preventing image distortion. Considering the overall dimensions of the bark removal equipment and the height of the installation site, the mechanical framework of the vision inspection system is designed to be 0.65 m (width) × 0.85 m (length) × 1.85 m (height). The vertical distance between the horizontal plane of the clamping fixture and the industrial camera is set at 0.62 m, ensuring that the camera can fully capture images of the Wood Panels with Bark within its field of view. The light source is positioned 0.56 m above the clamping fixture to provide appropriate illumination, thereby mitigating interference from ambient light and ensuring the quality of the captured images of the wood panels.

To address the issue of insufficient light transmission in the sawmill production environment, we selected a white light source and the Shenzhen Spit LED POWER 050A-2 strip industrial light, which has a luminous flux of 1150 lm, meeting the brightness requirements for image acquisition. Considering both quality and cost factors, we chose the ADVANTECH PPC-112W-PK92AA industrial panel computer as the model for the industrial control computer. This industrial computer is equipped with an embedded Rockchip 3399 processor, featuring a dual-core Cortex-A72 clocked at up to 1.8 GHz and a quad-core Cortex-A53 clocked at up to 1.4 GHz. It also has 2 GB of LPDDR4-1333 RAM and 16 GB of eMMC NAND flash storage, fulfilling the performance requirements for processing Wood Panels images. For the selection of the industrial camera and lens, we opted for the Dahua A3600CU60 color area scan industrial camera and the Canon EF-1806-5MP industrial lens. The industrial camera employs an IMX178 sensor, with a sampling frequency of 60 Hz and a resolution of 3072 × 2048 pixels, meeting the efficiency and clarity requirements for image acquisition. Fig. 2 illustrates the assembled vision inspection system within the bark removal equipment in the actual sawmill production environment.

The functions of the visual inspection system mainly include two parts: image acquisition and image processing, as shown in Figure 3. The image acquisition part utilizes an industrial camera to collect a large number of raw image data of wood panels, performing data augmentation operations to expand the quantity and types of the image set, making it more variable. Based on the shape, size, and position information of the bark-covered wood panels, annotations are made to construct a semantic segmentation dataset for training the bark removal semantic segmentation model. In the actual bark removal processing, the visual detection device is primarily used to acquire image data of the wood panels, enabling the visual detection system to perform semantic segmentation and feature parameter calculations. The image processing part involves two main tasks: on one hand, it conducts offline training of the semantic segmentation model using the constructed bark-covered wood panel semantic segmentation dataset. The semantic segmentation model is then used to extract features from the bark-covered wood panel images, perform semantic segmentation, and utilize evaluation metrics to validate and assess the segmentation performance of the model. On the other hand, the trained semantic segmentation model is packaged and used for wood panel image recognition to generate predicted masks. Edge detection techniques are then applied to identify the edges of the bark, fitting straight lines to the detected edges to calculate the cuttable width, posture adjustment values, and cutting channel values for the bark-covered wood panels. Finally, the model obtained from the training and testing is deployed to the bark removal equipment for practical bark removal work.

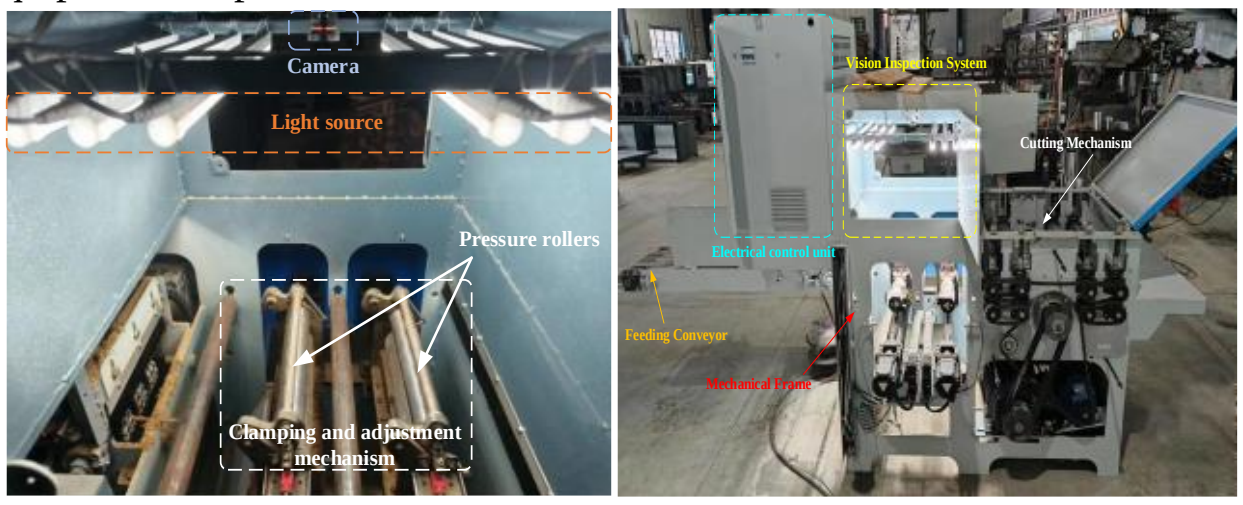

**Fig. 2. Customized visual inspection system and its installation results.**

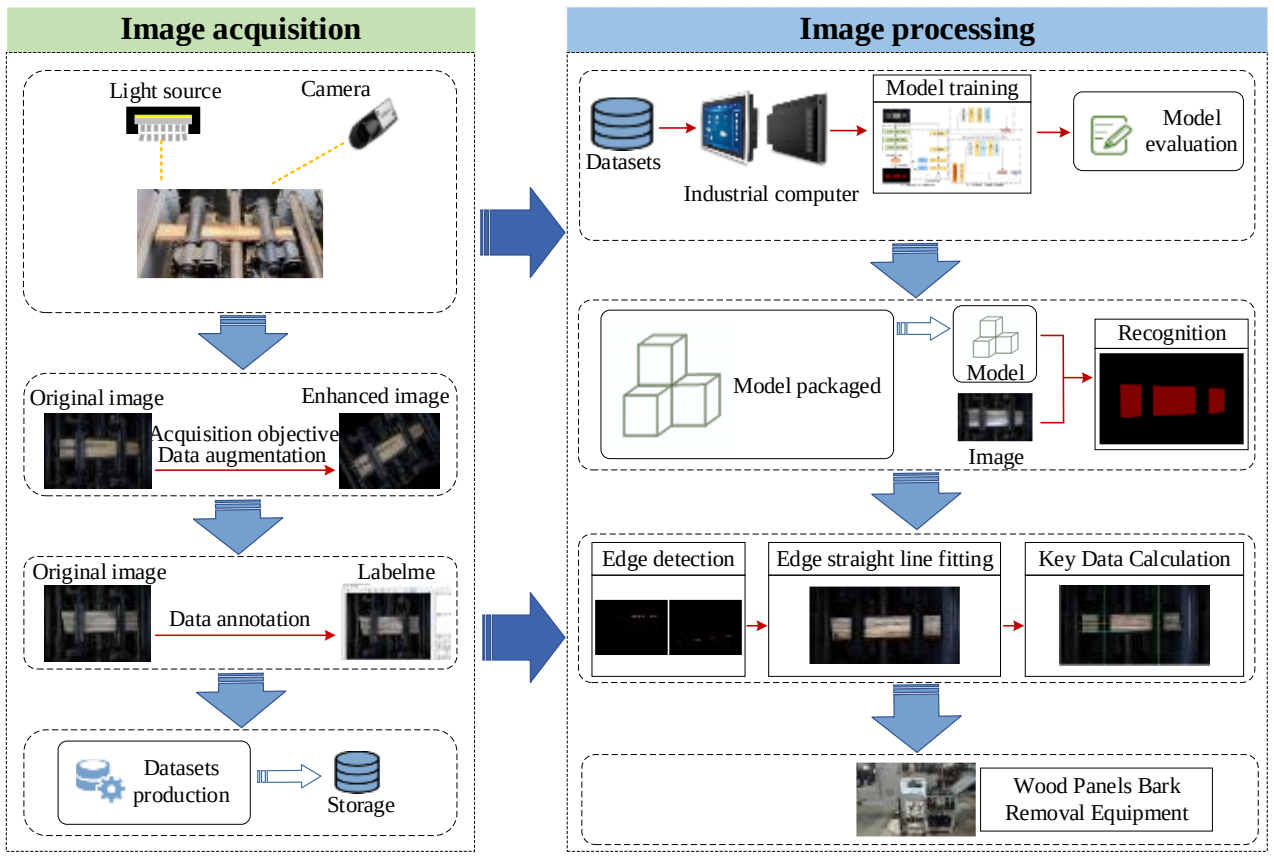

**Fig.3. Functions of the customized visual inspection system**

### *3.2. Semantic segmentation dataset for wood panels with bark*

Rich data is a crucial foundation for training deep learning network models. This paper constructs a comprehensive semantic segmentation image dataset of wood panels with bark in an industrial environment, following steps such as data collection, data augmentation, data annotation, and dataset recording and partitioning. This dataset is intended for subsequent offline training of the semantic segmentation model, while also providing data support for related research work.

During the bark removal process, the image data collection of wood panels with bark is accomplished through an industrial camera within the visual inspection system, allowing for the automatic acquisition of a rich set of image samples. When the wood panels with bark are positioned in the designated area of the panel holding and adjusting mechanism, the industrial camera captures the overall appearance of the panels, generating color images with a resolution of 3072 × 2048 pixels. We also utilize two external hard drives, each with a capacity of 512 GB, to collect and store the acquired images of wood panels with bark. In this study, the images can be categorized into panel areas and background areas (including bark), with their specific distribution depending on their shapes. Fig. 4 illustrates examples of the distribution of certain image areas.

To address the issue of overfitting and enhance the generalization ability of the model, we performed data augmentation on the images without altering their aspect ratios, ensuring that the features of the wood panels were fully retained. The augmentation techniques employed included rotation, horizontal flipping, and vertical mirroring. Through these data augmentation operations, we effectively increased both the quantity and diversity of images of wood panels with bark, thereby expanding the dataset.

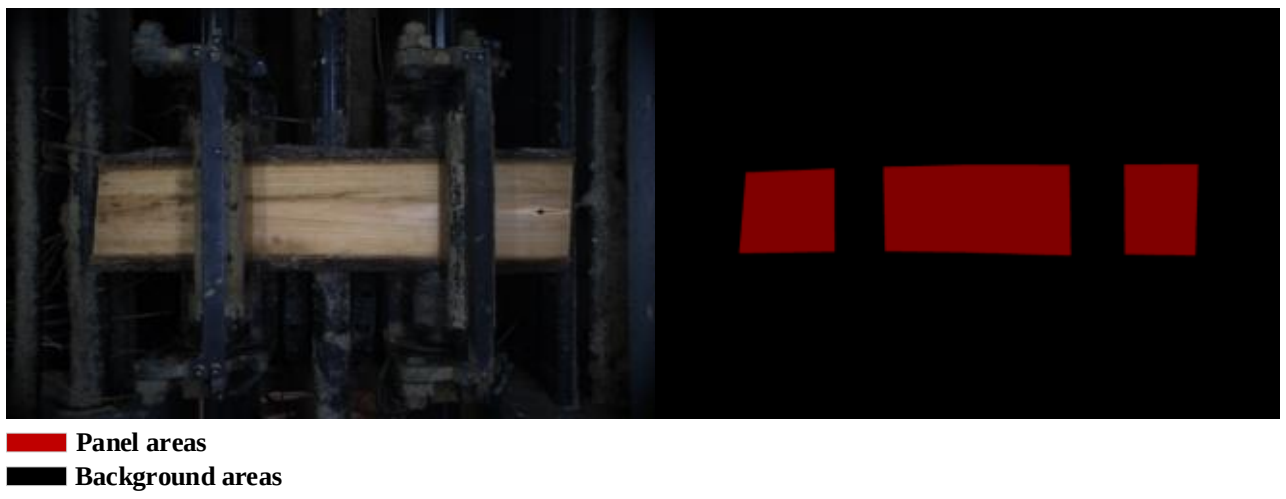

**Fig.4. Examples of the distribution of areas.**

The data annotation process primarily utilized Labelme [30]. To ensure the rigor of the annotation task, the dataset was annotated by two researchers. One researcher independently annotated the dataset in JSON format using Labelme, while the other researcher was responsible for verifying the annotation standards and correcting any non-standard image annotations. Fig. 5 illustrates partial examples of annotated images.

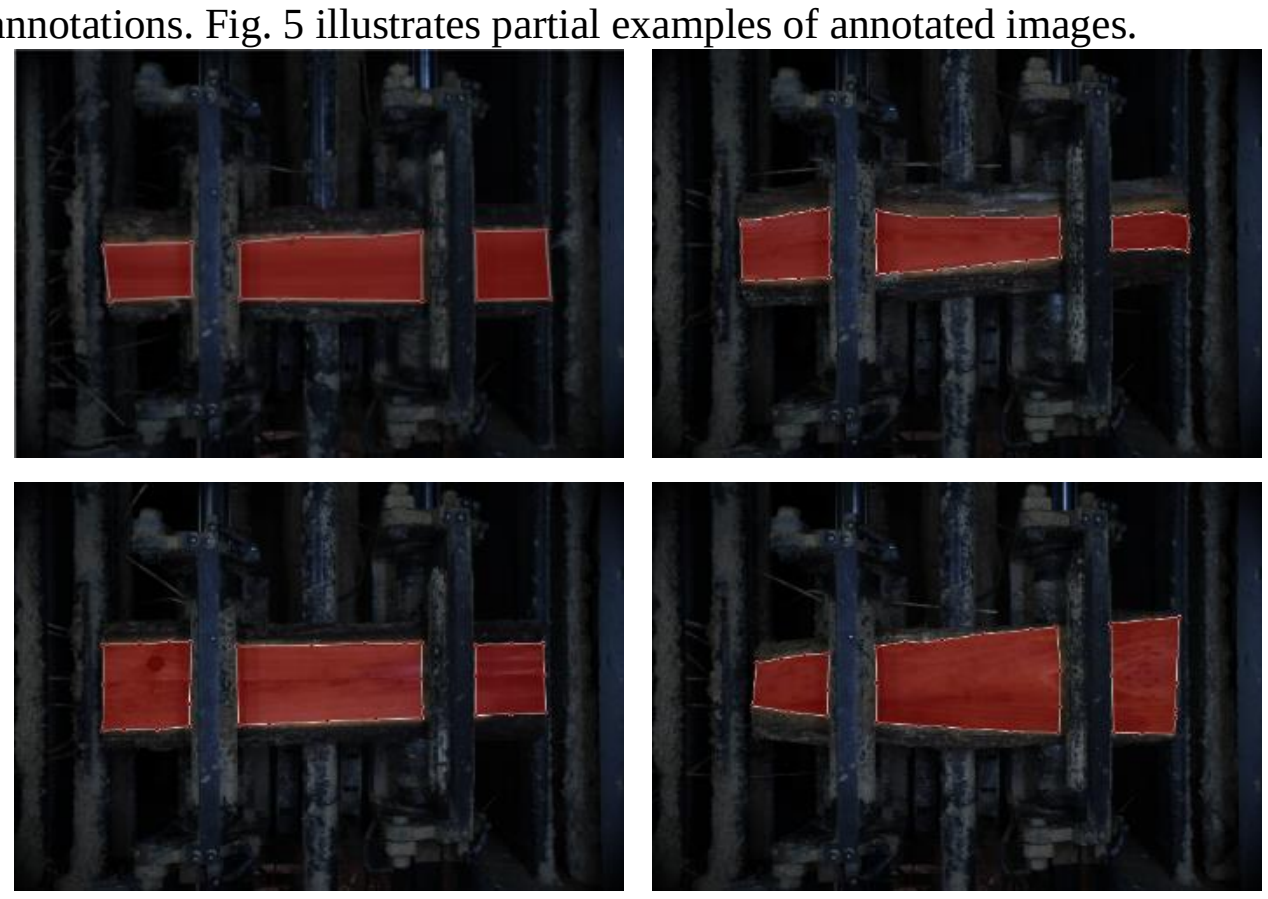

**Fig. 5. Partial examples of annotated images**

After completing the annotation process, we recorded and partitioned the constructed dataset. The dataset samples were collected from the designed visual inspection system, consisting of 6,484 images of wood panels with bark. Among these images, we randomly divided the dataset into training and validation sets in an 8:2 ratio, with 5,187 images designated for training and 1,297 images for validation. Both sets contain the original images of the wood panels along with their corresponding annotation information.

## 4. Semantic segmentation and key data calculation

### *4.1. Bark semantic segmentation based on BiSeNetV1*

This study employs the BiSeNetV1 [31] lightweight real-time semantic segmentation model, which balances computational efficiency and segmentation accuracy, to segment wood panels with bark. This model ensures both the accuracy and speed of segmentation, thereby meeting the sawmill's requirements for real-time performance and precision in bark removal processing. BiSeNetV1 is a dual-path segmentation network that features spatial and context paths. The network primarily consists of three components: the bilateral segmentation network architecture, the attention refinement module, and the feature fusion module. The process of utilizing BiSeNetV1 for semantic segmentation of wood panels with bark is illustrated in Fig. 6.

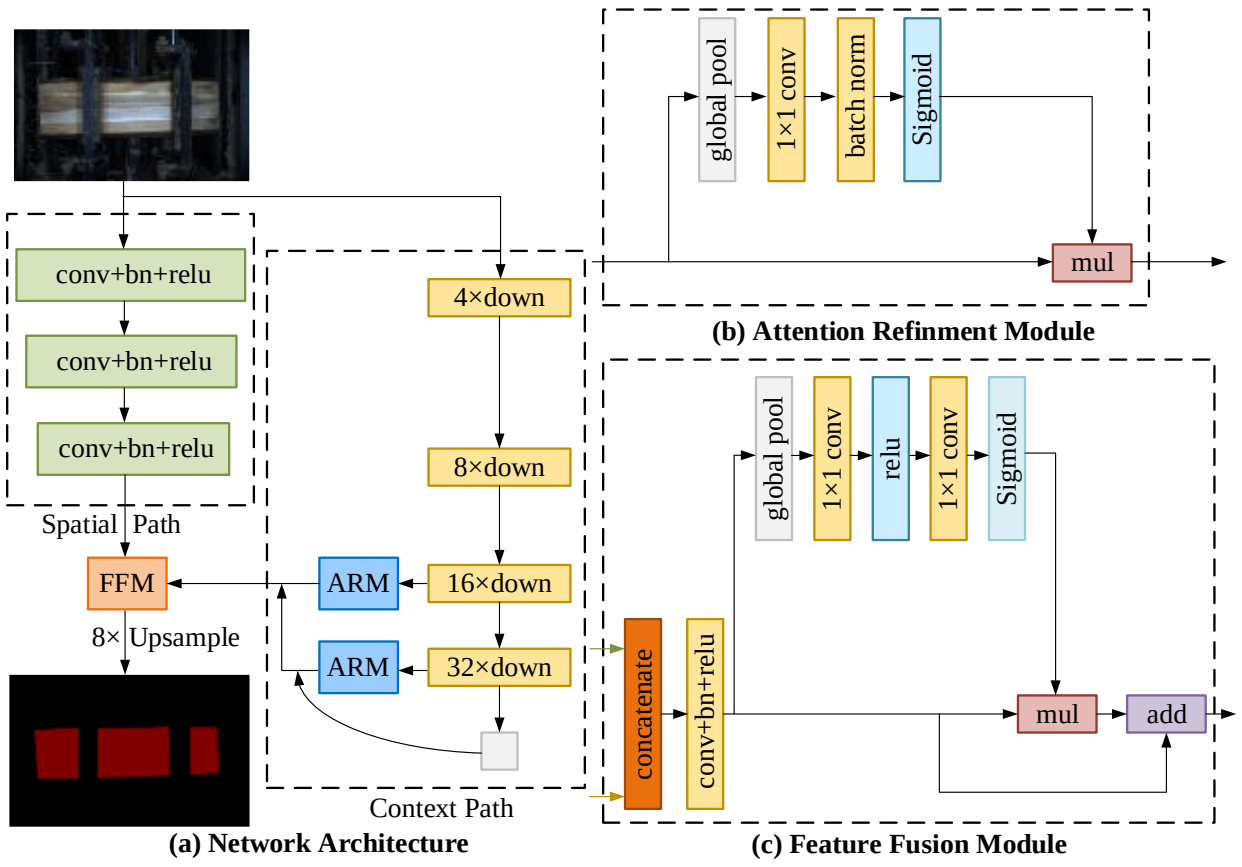

**Fig. 6. Processing of BiSeNetV1 for semantic segmentation**

The bilateral segmentation network architecture comprises two components: the Spatial Path and the Context Path. The Spatial Path consists of three convolutional layers, each followed by batch normalization and ReLU activation. Each convolutional layer has a stride of 2, resulting in output feature maps that are 1/8 the size of the original image of wood panels with bark, effectively preserving spatial information while enabling rapid responses. The Context Path utilizes a lightweight model and global average pooling to obtain the largest receptive field, thereby extracting global information to capture contextual features within the wood panel images. The pretrained ResNet18 model serves as the backbone for the Context Path, providing rich high-level semantic features. In the Context Path, the Attention Refinement Module (ARM) is employed to introduce an attention mechanism, enhancing the network's focus on useful features while reducing the emphasis on irrelevant information. This allows the network to concentrate more effectively on the segmentation task for the panel areas. Finally, the Feature Fusion Module (FFM) integrates features from both paths, achieving complementary feature fusion and capturing semantic information from the wood panel images more comprehensively. This integration improves segmentation accuracy without sacrificing speed, making the bark removal process more efficient and precise. Additionally, the BiSeNetV1 model employs a primary loss function to supervise the output of the entire model, along with two specific auxiliary loss functions to oversee the outputs of the Context Path, thereby providing a more intuitive assessment of the model's predictive performance.

To deploy the BiSeNetV1 model onto the designed wood panel bark removal equipment, this study utilizes the PyTorch framework for training the BiSeNetV1 model. The model is then exported in ONNX [33] format and deployed on the industrial computer, with performance optimization achieved through TensorRT to enable efficient inference capabilities. During the deployment process, it is necessary to install ONNX Runtime [34] on the industrial computer to support the inference of the ONNX model. To further enhance inference efficiency, TensorRT [35] and its related dependencies are installed. The *trtexec* tool is employed to convert the ONNX model into a TensorRT engine, utilizing NVIDIA's TensorRT for model optimization. This approach aims to improve inference performance and reduce memory usage.

### *4.2. Key data calculation*

To achieve bark removal after obtaining the panel areas using the semantic segmentation algorithm described in Section 4.1, it is essential to calculate the key data required by the electrical control unit. This key data includes the attitude of the wood panels (specifically, their main axis and angular orientation), the width available for cutting, and the distance to move to the corresponding cutting channel. The process for calculating the key data for bark removal is illustrated in Fig. 7.

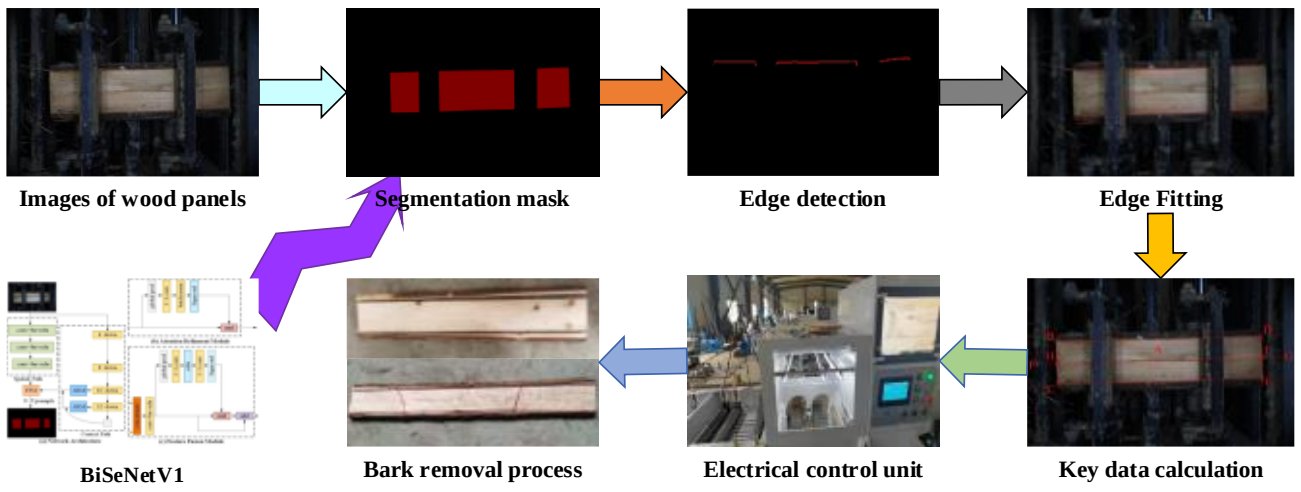

**Fig. 7. Process of key data calculation**

Images of wood panels are captured using an industrial camera, and the segmentation mask is generated using the BiSeNetV1 model. Edge detection operators are designed to obtain the upper and lower edges of the wood panel in the image. The least squares method, combined with the Tukey approach, is employed to fit these two edges as straight lines. By calculating the necessary parameters, the key data for bark removal are obtained and sent to the electrical control unit to drive the actuators, thereby completing the bark removal process.

#### *4.2.1. Edge detection based on the Prewitt operator*

In order to calculate the width available for cutting, we first need to identify the upper and lower edges of the wood panel. The segmented mask obtained from the semantic segmentation model clearly defines the transverse edge between the bark and the panel areas. The edge information is prominent, the background is simple, and the noise level is low. Given these characteristics, this paper employs the commonly used Prewitt operator in image processing to detect the upper and lower edges of the wood panel. The Prewitt operator is a convolution-based edge detection method designed to detect edge information in images while offering a certain degree of noise suppression. During edge detection, the operator applies horizontal and vertical kernels to the image through convolution operations. The results from these operations are squared, summed, and then square-rooted to produce the final edge strength. Noise suppression is achieved by averaging pixel values, which functions similarly to a low-pass filter[21]. The specific steps are as follows:

**Step 1:** Determine the size of the convolution kernels. Smaller convolution kernels are computationally efficient and sensitive to local details. Therefore, a 3×3 kernel is chosen to maintain computational efficiency while being sensitive to fine structures and small-scale features in the image.

**Step 2:** Construct the convolution kernels. Two vertical convolution kernels, $G_{Y1}$ and $G_{Y2}$, are constructed as shown in Equation (1). $G_{Y1}$ is used to detect edges transitioning from light to dark, while $G_{Y2}$ detects edges transitioning from dark to light.

**Step 3:** Apply the convolution kernel $G_{Y1}$ to the image to perform convolution operations, detecting the upper edge of the wood panel.

**Step 4:** Apply the convolution kernel $G_{Y2}$ to the image to perform convolution operations, detecting the lower edge of the wood panel.

$$G_{Y1} = \begin{bmatrix} 1 & 1 & 1 \\ 0 & 0 & 0 \\ -1 & -1 & -1 \end{bmatrix} \quad G_{Y2} = \begin{bmatrix} -1 & -1 & -1 \\ 0 & 0 & 0 \\ 1 & 1 & 1 \end{bmatrix} \tag{1}$$

After performing the above steps to calculate the edges, we observed that the detected edges are often short and discontinuous, which significantly affects the subsequent step of fitting straight lines to the wood panel edges. To reduce this impact, we filter out the shorter and discontinuous edges. The final edge detection result for the wood panel areas is shown in Fig. 8.

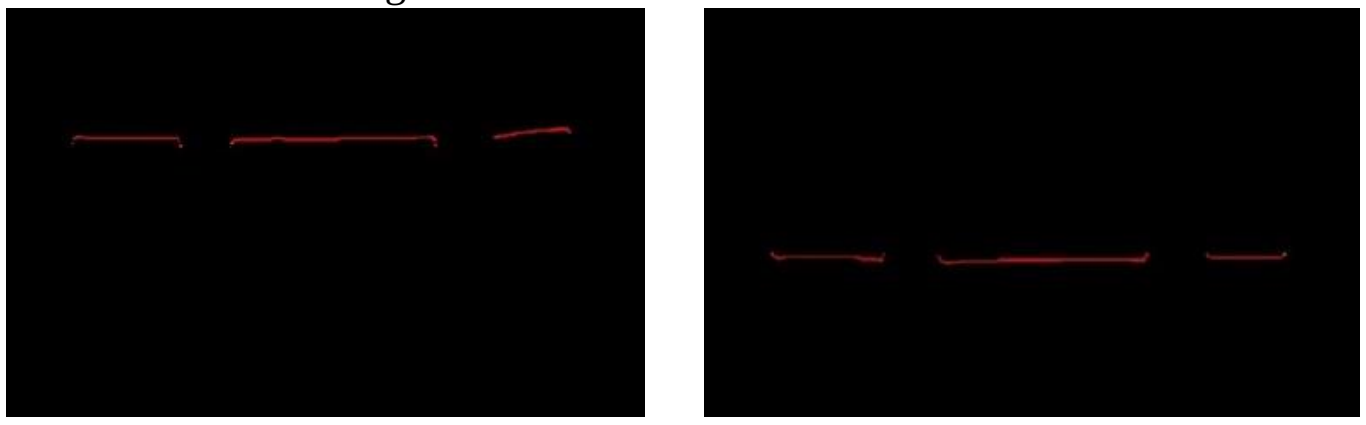

**(a) upper edges** **(b) lower edges**

**Fig. 8 An example of edge detection results.**

### *4.2.2. Edge Fitting of the wood panel areas*

For some wood panel images with relatively regular edges, edge detection algorithms can effectively identify continuous wood panel edges. However, for wood panels with highly irregular or fluctuating edges, the edge detection method based on the Prewitt operator struggles to cope, which presents challenges for edge fitting of the panel area. To address this issue, this paper leverages the robustness of the Tukey algorithm by combining the Tukey concept with the least squares method for linear edge fitting of the wood panel images. This combination allows for more robust and accurate edge line fitting, especially in handling noise and outliers. The specific steps are as follows:

**Step 1:** Initial Fitting. Using the edge data points of the wood panel area obtained in Section 4.2.1, which are typically represented as a set of 2D points ($x_i$, $y_i$), where $1 \le i \le N$, the least squares method is applied to perform an initial fit. This process results in the preliminary linear parameters: the slope $k$ and the intercept $b$.

$$k = \frac{N \sum x_i y_i - \sum x_i \sum y_i}{N \sum y_i - (\sum x_i)^2}$$

$$b = \frac{\sum y_i - k \sum x_i}{N} \tag{2}$$

**Step 2:** Compute Initial Residuals. Based on the initial fitting results from Step 1, calculate the residual $r_i$ for each point, which represents the distance between each data point and the fitted line.

$$r_i = y_i - (kx_i + b) \tag{3}$$

**Step 3:** Apply Tukey's Loss Function. For each residual $r_i$, apply Tukey's loss function to calculate the weight. Set a threshold $c$, and for each residual, determine whether its absolute value is less than $c$:

If $|r_i| < c$, the weight can be calculated using Tukey's loss function formula as follows:

$$\omega_i = 1 - (1 - \frac{|r_i|}{c})^2 \tag{4}$$

If $|r_i| \ge c$, then the weight $\omega_i = 0$, meaning these points no longer influence the fitting process. This approach helps to reduce the impact of outliers and noise, allowing for a more robust linear fit for the wood panel edges.

**Step 4:** Utilize the calculated weights $\omega_i$ to perform weighted least squares regression, which involves recalculating the slope $k$ and intercept $b$.

$$k = \frac{\sum \omega_i (x_i - \bar{x})(y_i - \bar{y})}{\sum \omega_i (x_i - \bar{x})^2}$$

$$b = \bar{y} - k\bar{x} \tag{5}$$

Where $\bar{x}$ and $\bar{y}$ are the weighted averages, and $\bar{x} = \frac{\sum \omega_i x_i}{\sum \omega_i}$, $\bar{y} = \frac{\sum \omega_i y_i}{\sum \omega_i}$.

**Step 5:** Iterative Update. The aforementioned steps can be repeated to calculate new residuals and weights until convergence is achieved, which is defined as when the parameter changes are less than a specified threshold. The updated parameters, namely the slope $k$ and intercept $b$, are then obtained.

Through these steps, the fitted edges of the wood panel areas can better adapt to the edge data points, thereby reducing the impact of outliers and enhancing the fitting accuracy. Fig. 9 presents partial examples of the edge line fitting results for the wood panel images.

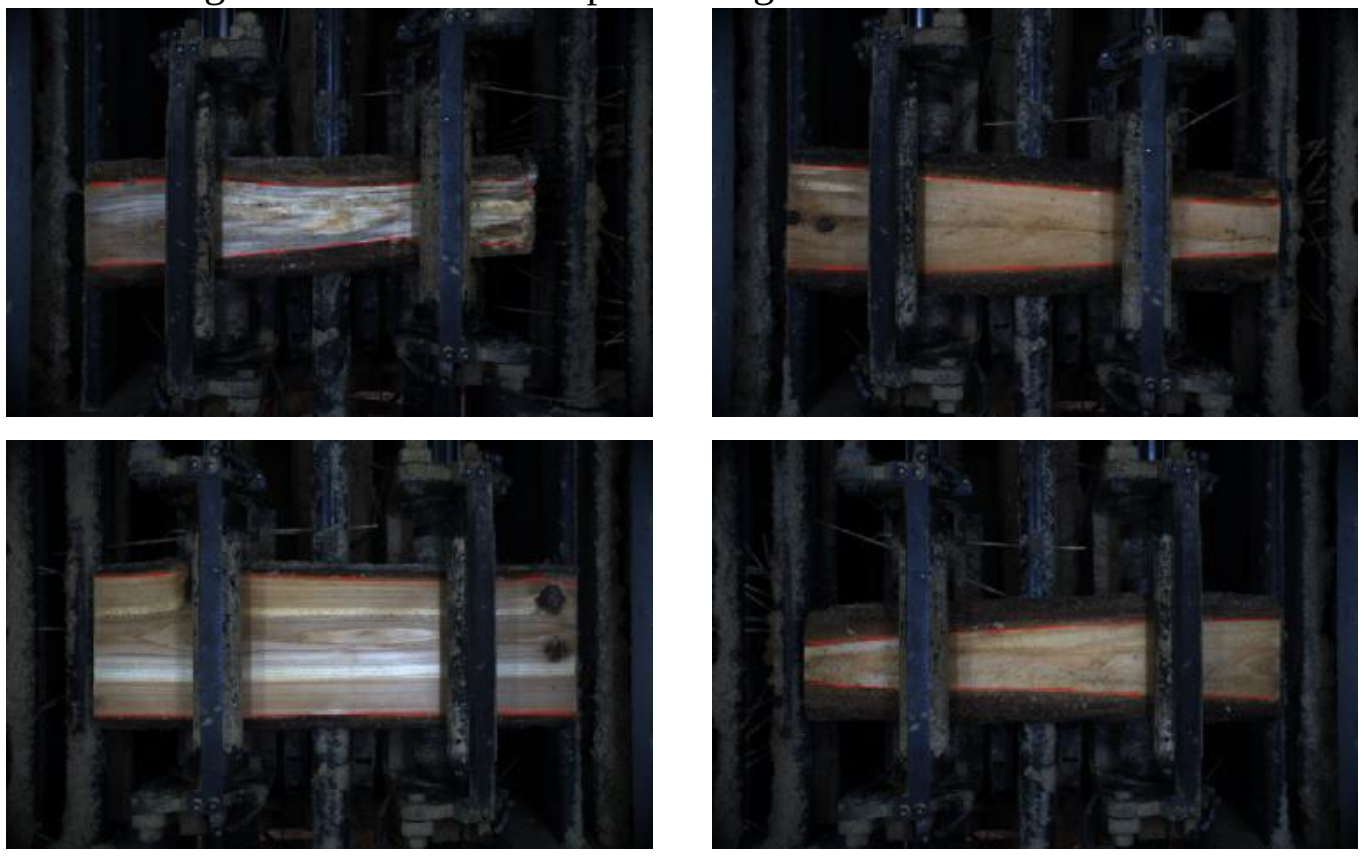

**Fig. 9 Partial examples of the edge line fitting results.**

### *4.2.3 Key data calculation for panel areas*

The calculation of key data for panel areas is essential for guiding the electrical control unit to accurately execute the bark removal task. Given that each wood panel varies in width and that the attitude (specifically, the main axis and angular orientation) differs upon entering the field of view of the industrial camera, the distance to move to the corresponding cutting channel—referred to as the cutting distance—remains uncertain due to variations in the available width for cutting. Consequently, precise calculation of these key data is critical for achieving accurate bark removal.

(1) Attitude of the wood panels

Due to the varying orientations of the wood panels upon reaching the field of view of the industrial camera, it is essential to conduct an attitude calculation to determine the main axis and angular orientation of the target wood panel. This calculation subsequently allows for the determination of two key data points: the width available for cutting and the cutting distance. These data are crucial for achieving precise bark removal. Fig.10 illustrates the schematic of the attitude data calculation for wood panels.

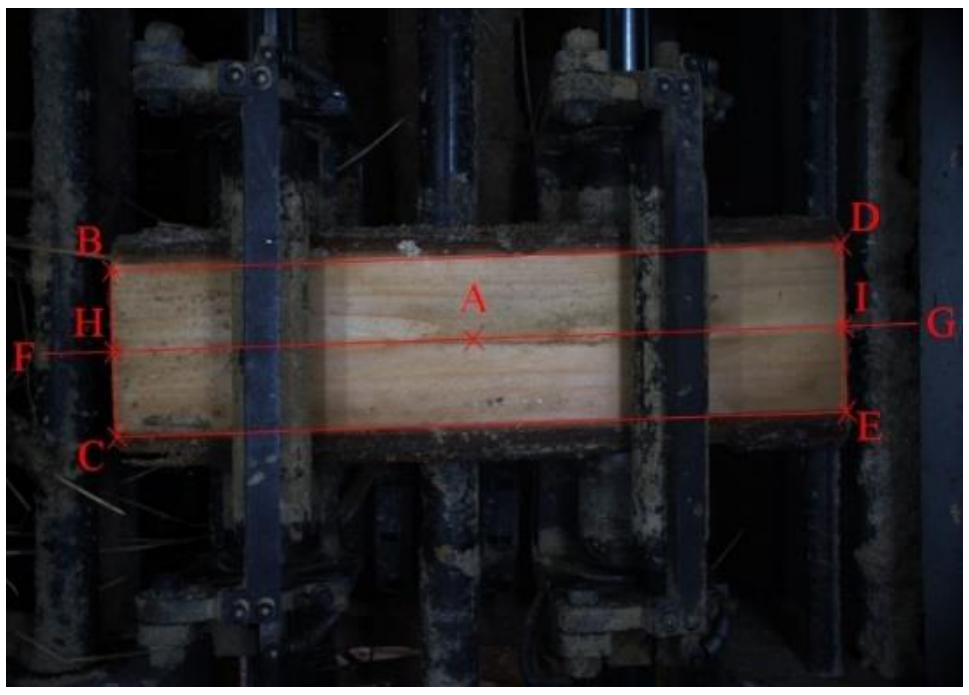

**Fig. 10. Schematic of the attitude calculation.**

The steps involved in the calculations and transformations are as follows:

**Step 1:** Calculate the Centroid Coordinates of the wood panel A($x_o$, $y_o$)

**Step 2:** Traverse the Endpoints of the Fitted Edge Lines. Traverse the endpoints of the fitted edge lines to obtain the coordinates of the upper left corner point B($x_b$, $y_b$), the lower left corner point C($x_c$, $y_c$), the upper right corner point D($x_d$, $y_d$), and the lower right corner point E($x_e$, $y_e$).

**Step 3:** Connect Points to Form Lines. Connect point Band point C to form a line segment BC, and connect point D and point E to form another line segment DE.

**Step 4:** Calculate the Second-Order Central Moments. Calculate the second-order central moments of the panel area, denoted as ($M_{11}$, $M_{20}$, $M_{02}$), where,

$$M_{ij} = \sum_{(x,y)\in R}(x - x_o)^i\,(y - y_o)^j \qquad (6)$$

In the above equations, $M_{ij}$ represents the central moments, where $i$ and $j$ denote the orders of the moments, with $i$ corresponding to the horizontal coordinate and $j$ corresponding to the vertical coordinate. The coordinates ($x$ , $y$) refer to the pixel locations within the panel area.

**Step 5:** Utilize equation (6) to calculate the angular orientation $\theta$ of the wood panel, thereby determining the main axis and the angular orientation of the wood panel.

$$\theta = \frac{1}{2}tan^{-1}\left(\frac{2M_{11}}{M_{20}-M_{02}}\right) \qquad \mathbf{(7)}$$

(2) Calculation of the width available for cutting

Considering the irregularity of the panel edges, the width available for cutting can be defined as the minimum vertical distances from the upper edge point set (on line BC) and the lower edge point set (on line CE) to the wood panel's main axis reference line (line HI).

By iterating through the vertical distances from both the upper edge point set and the lower edge point set to the main axis reference line, we identify the minimum distances $w_{\text{minup}}$ and $w_{\text{mindown}}$ , which correspond to the points  and, respectively. Consequently, the width available for cutting is given by the sum of these two minimum distances:

$$w = w_{\text{minup}} + w_{\text{mindown}} \qquad (8)$$

(3) Cutting distance

To facilitate the calculation of the cutting distance, it is essential to determine the cutting centerline. The direction of the cutting centerline must align with the reference line of the wood panel's main axis, while also ensuring that the $w_{\text{minup}}$ and $w_{\text{mindown}}$ are equal. The calculation steps are as follows:

**Step 1:** Definition of the translation affine transformation matrix:

$$\begin{bmatrix} 1 & 0 & t_x \\ 0 & 1 & 0 \\ 0 & 0 & 1 \end{bmatrix}$$

**Step 2**: Connect points $u$ and $d$ to obtain line segment $ud$, and calculate its midpoint $m(x_m, y_m)$.

**Step 3:** Calculate the Translation Amount $t_x$. A vertical translation reference line N is drawn through point $m(x_m, y_m)$, parallel to the $y$-axis. The intersection point of the wood panel's main axis reference line HI and the translation reference line is denoted as $n(x_n, y_n)$. The translation amount $t_x$ is then calculated as the difference between the y-coordinate of point $m$ and the y-coordinate of point $n$.

**Step 4:** Using the affine transformation matrix and the translation amount $t_x$, the wood panel's main axis reference line HI is translated to obtain a new line L, which represents the cutting midline of the wood panel.

**Step 5:** Identify the midpoint $o(x_o, y_o)$ of line L and calculate the pixel length of line L. This allows for the generation of the cuttable rectangular area of the wood panel.

The cuttable centerline of the wood panel, along with the positional information of the saw channel centerline within the visual inspection area and the starting position of the wood panel, allows for the calculation of the cutting distance. The schematic diagram for calculating the cutting distance that the wood panel needs to move is shown in Fig. 11.

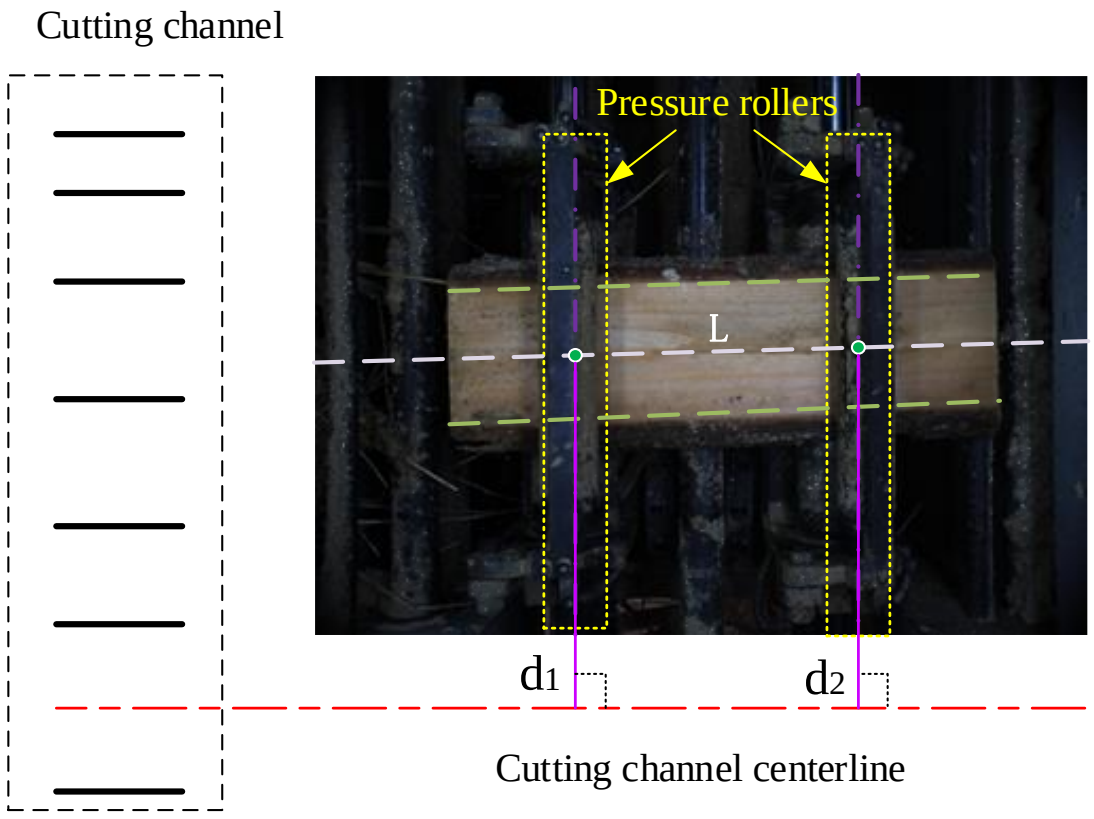

**Fig. 11 Schematic diagram for calculating the cutting distance**

The positional information of the cutting channel centerline in the visual inspection area can be derived from the position of the installed saw blades on the cutting channel and the spacing between the saw blades. The positions of the two pressure rollers in the wood clamping and adjustment mechanism are fixed within the visual inspection area. By calculating the intersection points between the pressure rollers and the cutting centerline, the initial position of the wood panel within the visual inspection area can be determined. Based on the initial position of the wood panel and the position of the cutting channel centerline, the cutting distances d1 and d2 that the wood panel needs to move can be calculated.

## 5. Experimental results and analysis

Under laboratory conditions, a comparative experiment was conducted with commonly used segmentation models to validate the effectiveness of the wood panel bark semantic segmentation model based on BiSeNetV1. Additionally, practical experiments for bark removal were carried out in a sawmill to test and verify the performance of the designed bark removal equipment.

### *5.1. Experiments under laboratory conditions*

#### *5.1.1 Training environment*

The configuration of the experimental environment is shown in Table 1.

**Table 1. Experimental environment configuration.**

| Configuration environment | Configuration name (version) |
|---|---|
| Operating system | Window 10 |
| CPU | 9th Gen Intel(R) Core(TM) i5-9300H |
| GPU | NVIDIA GeForce GTX 1650 |
| Compilers | Python 3.8.18 |
| Deep Learning Framework | Pytorch 1.10.0 |
| Acceleration Module | CUDA 10.2 |

The experimental hyperparameters were set as shown in Table 2.

**Table 2. Experimental environment configuration.**

| Hyperparameters | Values |
|---|---|
| Image size | 512 |
| Batch size | 4 |
| Iteration times | 100 |
| Learning rate | 0.0001 |
| Momentum | 0.9 |
| Weight_decay | 0.0001 |

#### *5.1.2 Evaluation criteria*

The performance of the model is evaluated from two aspects: segmentation accuracy and segmentation efficiency. Regarding the segmentation accuracy, the evaluation primarily utilizes the Mean Intersection over Union (MIoU) [36] and the Mean Pixel Accuracy (MPA) [36] to assess the model's segmentation precision.

The MIoU measures the average intersection over the union across all classes and can be described as follows:

$$\mathrm{MIoU}=\frac{1}{\mathrm{k}+1}\sum_{\mathrm{i}=0}^{\mathrm{k}}\frac{\mathrm{p_{ii}}}{\sum_{\mathrm{j}=0}^{\mathrm{k}}\mathrm{p_{ij}}+\sum_{\mathrm{j}=0}^{\mathrm{k}}\mathrm{p_{ji}}-\mathrm{p_{ii}}} \tag{9}$$

The MPA reflects the average classification accuracy of the model across all categories and can be described as follows:

$$\mathrm{MPA}=\frac{1}{\mathrm{k}+1}\sum_{\mathrm{i}=0}^{\mathrm{k}}\frac{\mathrm{p_{ii}}}{\sum_{\mathrm{j}=0}^{\mathrm{k}}\mathrm{p_{ij}}} \tag{10}$$

In equations (9) and (10), $k$ denotes the number of classes, while $k$+1 signifies the total number of classes, including the background. The variable $i$ corresponds to a specific class. It is assumed that the dataset comprises $k$+1 classes (0, ..., $k$), with $k$=0 typically representing the background. In this study, the value of $k$ is set to 1; $P_{ii}$ indicates the number of pixels within the board area that are correctly classified, $P_{ij}$ represents the number of pixels from the background that are misclassified as belonging to the board area, and $P_{ji}$ denotes the number of pixels from the board area that are misclassified as background.

For segmentation efficiency, the evaluation is based on two key aspects: model complexity and inference speed. Model complexity is primarily assessed using the number of Parameters (Params) and Giga Floating Point Operations Per Second (GFLOPs). The Params metric refers to the number of parameters that the model needs to learn, while the GFLOPs metric evaluates the number of floating-point calculations required by the model, providing insight into the model's real-time segmentation performance. In terms of inference speed, the model's efficiency is measured by Frames Per Second (FPS), which calculates the number of prediction images inferred by the model per second.

#### *5.1.3 Comparison of segmentation models*

In the comparative experiment, we aimed to objectively evaluate the segmentation accuracy and efficiency of the BiSeNetV1 by comparing it with established semantic segmentation models, including PSPNet [37], ICNet [38], UNet [39], HRNetV2 [40], and DeepLabV3+ [41]. The results of these experiments are presented in Tables 3 and 4.

**Table 3. Comparison results of segmentation accuracy.**

| Network model | MIoU/% | MPA/% |
|---|---|---|
| PSPNet | 90.73 | 97.76 |
| ICNet | 97.40 | 99.39 |
| UNet | 98.24 | 99.60 |
| HRNetV2 | 97.87 | 99.52 |
| DeepLabV3+ | 97.29 | 99.39 |
| BiSeNetV1 | **97.67** | **99.48** |

**Table 4. Comparison results of segmentation efficiency.**

| Network model | Params/M | GFLOPs | FPS/fps |
|---|---|---|---|
| PSPNet | 2.38 | 6.03 | 54.08 |
| ICNet | 26.24 | 73.93 | 21.56 |
| UNet | 24.89 | 451.67 | 6.76 |
| HRNetV2 | 9.64 | 37.32 | 16.98 |
| DeepLabV3+ | 5.81 | 52.87 | 35.03 |
| BiSeNetV1 | **12.80** | **26.09** | **61.87** |

The combined analysis of Tables 3 and 4 indicates that BiSeNetV1 does not outperform other segmentation models in terms of both segmentation accuracy and efficiency. For example, regarding segmentation accuracy, Unet exhibits the best performance, achieving an MIoU of 98.24% and an MPA of 99.60%. In terms of efficiency, PSPNet demonstrates the lowest Params and GFLOPs, indicating high operational efficiency. However, while Unet excels in accuracy, its efficiency is comparatively lower, particularly in GFLOPs and FPS metrics, which significantly restricts its applicability in industrial bark removal tasks.

Conversely, although PSPNet achieves the highest efficiency, its accuracy performance is the weakest, which may lead to inaccurate bark removal in practical applications, thereby compromising product quality and resulting in timber resource waste. Notably, BiSeNetV1 achieves a commendable balance between segmentation accuracy and efficiency, thus providing optimal overall performance. This balance is essential for bark removal equipment used in sawmill environments, where both precision and production efficiency are critical. Therefore, the application of the BiSeNetV1 segmentation model in bark removal equipment is both rational and feasible.

To visually demonstrate the segmentation effects of the aforementioned models on wood panel images, a comparative analysis of the predicted masks and Ground Truth was conducted using visualization techniques. This analysis focused on the differences in edge information, correctly predicted regions, and detail handling among the models. Four groups of wood panel images with varying shapes were selected for visualization, as shown in Fig.12.

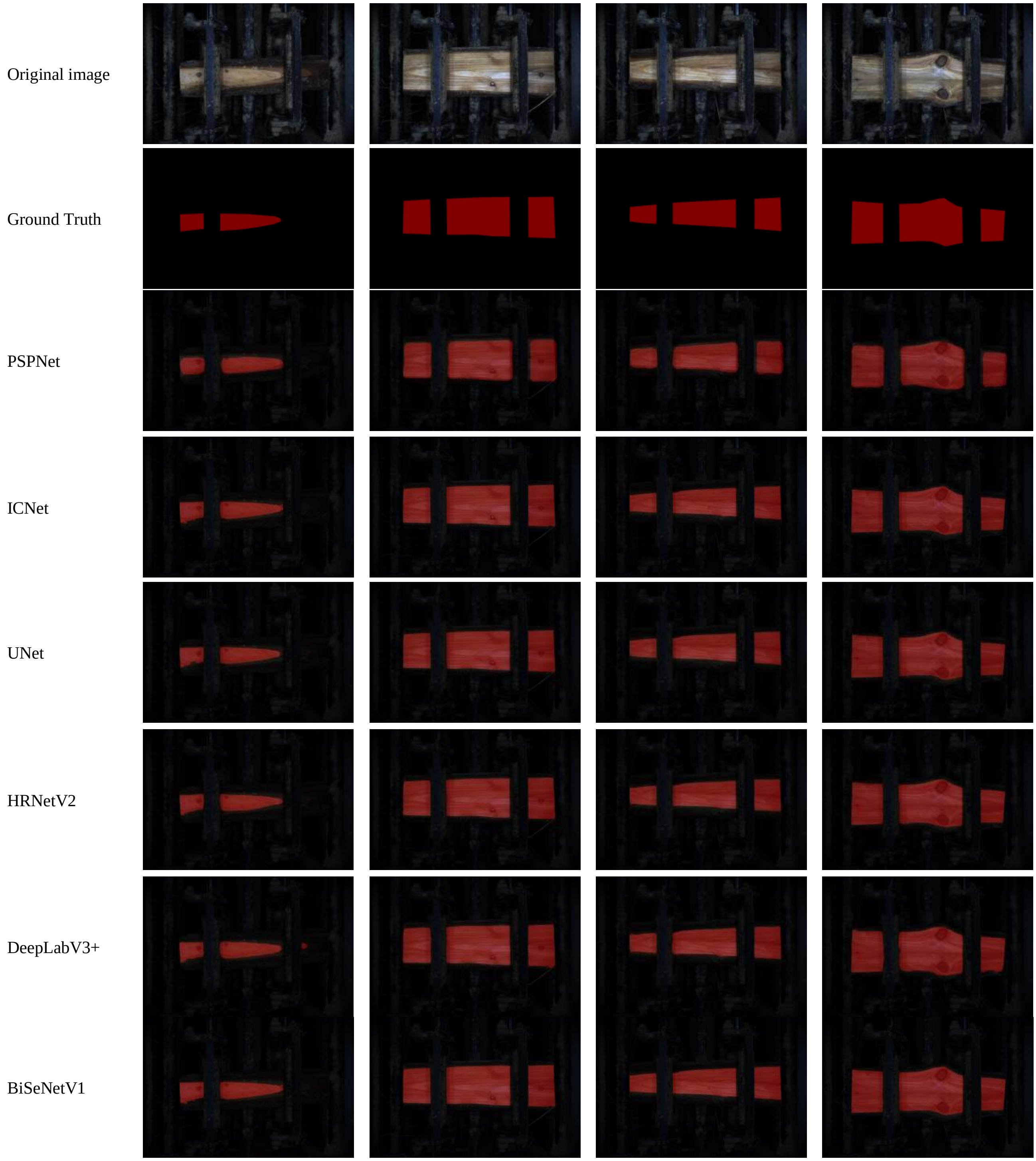

**Fig. 12. Partial visualization results.**

It is evident that while PSPNet captures the general segmentation contours of the wood panels, its predicted results are relatively coarse compared to Ground Truth. Notably, the edge processing of PSPNet's masks is quite blurred, failing to accurately capture the boundary details of the wood panels. In contrast, UNet successfully delineates the main contours of the wood panels, with smoother edge predictions that closely resemble the true outlines, exhibiting better detail handling overall and minimal discrepancy from Ground Truth. These observations align with the conclusions drawn from the performance metrics. The masks predicted by the BiSeNetV1 model closely match Ground Truth, with fine detail handling across different regions, effectively preserving object details while also achieving efficient separation of the wood panel areas in images with irregular shapes.

### *5.2. Wood bark removal experiments in a sawmill*

To verify whether the bark removal equipment meets the actual production requirements of sawmills, we conducted validation and effectiveness testing experiments for bark removal under factory conditions. The entire experiment was divided into two steps. In the first step, we utilized wood panels that had already undergone bark removal (referred to as standard wood panels) to conduct a validation experiment on the designed equipment, confirming the accuracy of the semantic segmentation model in identifying and segmenting the wood panel areas. In the second step, we performed effectiveness testing on bark removal using wood panels with bark. This step aimed to validate the accuracy of the critical data calculations for bark removal while assessing the effectiveness of the designed equipment in the context of the actual production requirements of the sawmill.

#### *5.2.1 Bark removal validation experiment*

We selected four groups of commonly used standard wood panels of varying sizes, as illustrated in Fig. 13. The nominal widths of the panels were 4.20 cm, 5.20 cm, 6.20 cm, and 7.20 cm, with a total of 40 pieces in each group. Each panel was tested (to prevent waste, no saw blades were installed in the cutting channel, as shown in Fig. 14), and each group underwent two repetitions of testing. The average values from each test were recorded for comparison with the nominal widths. The experimental results are summarized in Table 5. The maximum error of the repeated measurement values does not exceed 0.05cm, while the error of the first and second measurements is within 0.15 cm compared to the nominal width of the standard wood panels. These results demonstrate that the segmentation model employed in this study is capable of accurately identifying and segmenting the wood panel areas.

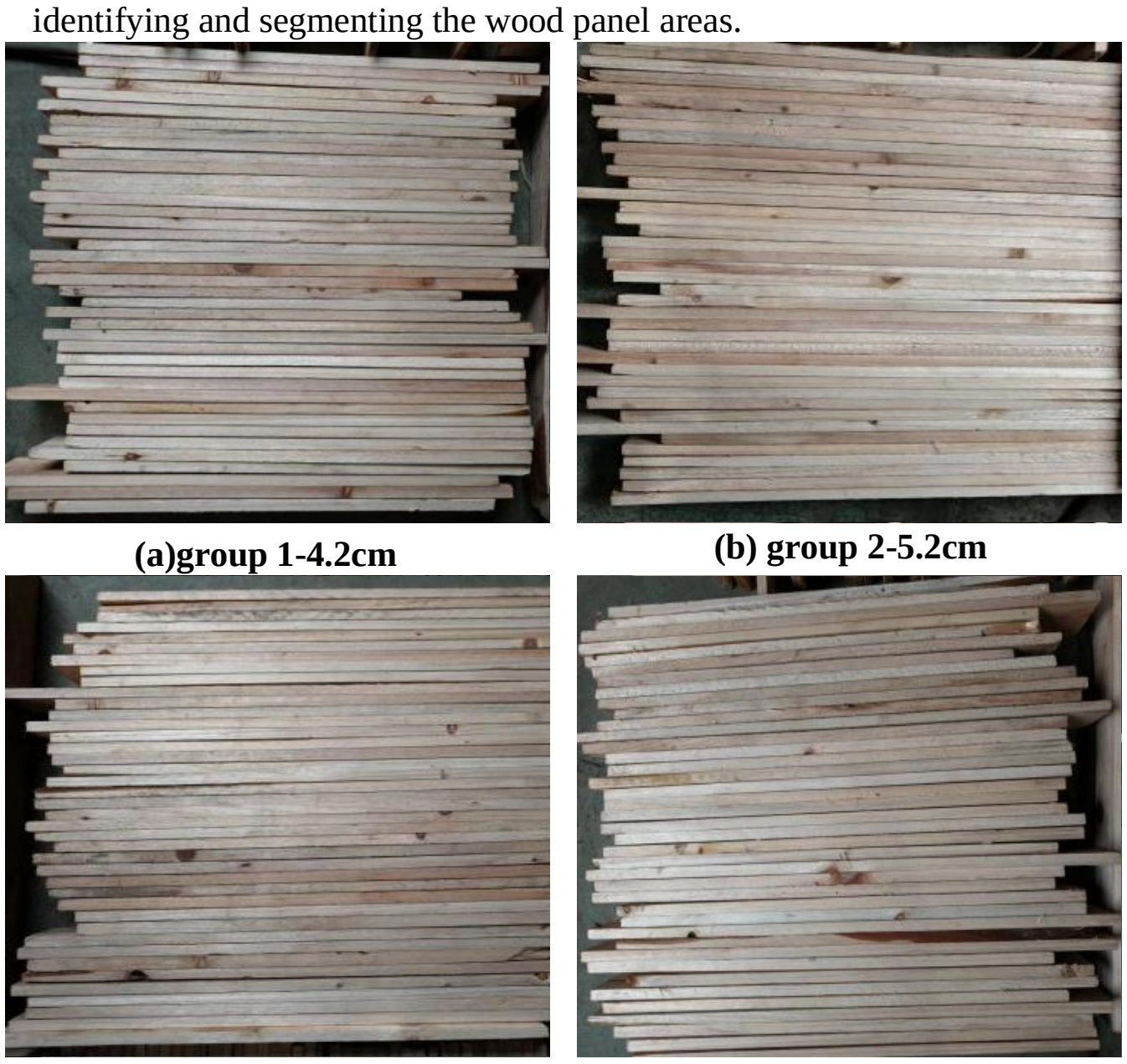

**(a)group 1-4.2cm** **(b) group 2-5.2cm**

**(c) group 3-6.2cm** **(d) group 4-7.2cm**

**Fig. 13. Four groups of standard wood panels.**

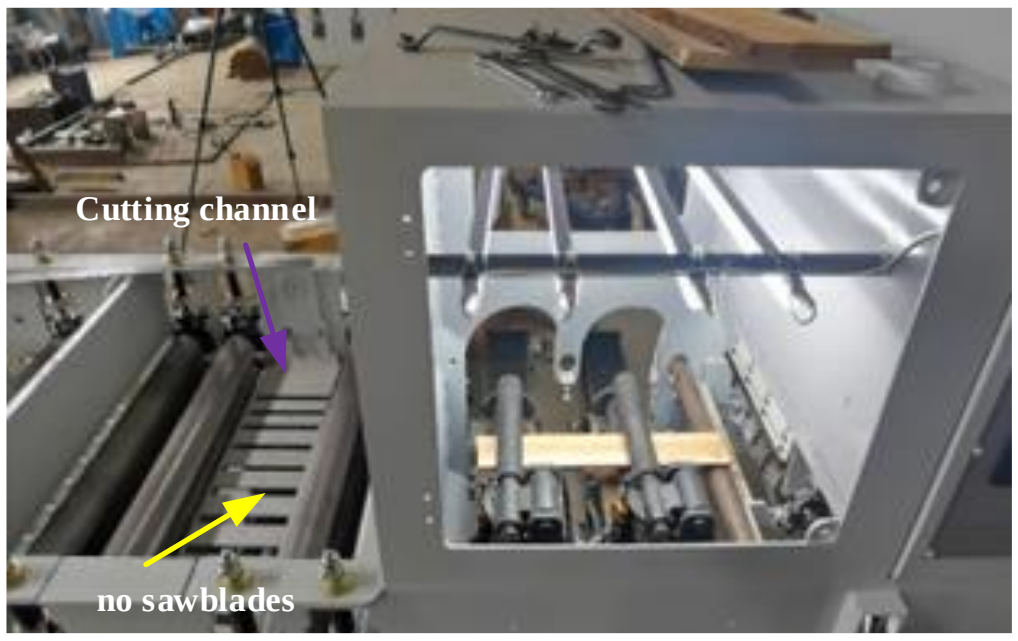

**Fig. 15. Equipment for bark removal validation experiment**

**Table 5. Results of bark removal validation experiment.**

| Group \ Time | Average of first (cm) | Average of second (cm) | nominal width (cm) |
|---|---|---|---|
| 1 | 4.12 | 4.11 | 4.20 |
| 2 | 5.09 | 5.06 | 5.20 |
| 3 | 6.07 | 6.11 | 6.20 |
| 4 | 7.14 | 7.11 | 7.20 |

#### *5.2.2 Wood bark removal effectiveness testing experiment*

In the sawmill, testing Experiment were conducted using the developed wood bark removal equipment on wood panels with bark to evaluate whether the equipment meets the sawmill's requirements for precision and efficiency in bark removal processing. To align with the prolonged working conditions of traditional bark removal equipment operated by workers, a comparative experiment was conducted for one hour between the conventional equipment and the designed bark removal device. After the experiment, data recorded by the developed bark removal equipment (mainly the number of wood panels and the quantity of accurately removed bark) were exported. The data for the traditional equipment were calculated by experienced technicians at the sawmill, resulting in the test outcomes shown in Tables 6 and 7. From these tables, it can be seen that the designed bark removal equipment achieved a cutting accuracy rate of 97.72%, a cutting speed of 30 panels per minute, and an overall cutting efficiency of 1,814 panels per hour. These metrics significantly surpass manual cutting, fully satisfying the sawmill's demands for precision and efficiency in bark removal processing. Fig. 15 illustrates examples of the bark removal results, demonstrating that the bark was accurately removed and validating the accuracy of the key data calculations.

**Table 6. Test results of cutting accuracy under production conditions.**

| method | Total number /Piece | Correct number /Piece | Correct Rate /% |
|---|---|---|---|
| Hand cutting | 937 | 855 | 91.25 |
| ours | **1856** | **1814** | **97.72** |

Note: The number of correct cuts refers to the cut width of off-specification wood boards less than 1 cm after cutting the two edges together. The rate of correct cuts refers to the ratio of the number of correct cuts to the number of presidential cuts.

**Table 7. Test results of cutting efficiency under production conditions.**

| method | Cutting speed （Piece/minute） | Cutting efficiency （Piece/hour） |
|---|---|---|
| Hand cutting | 14 | 855 |
| ours | **30** | **1814** |

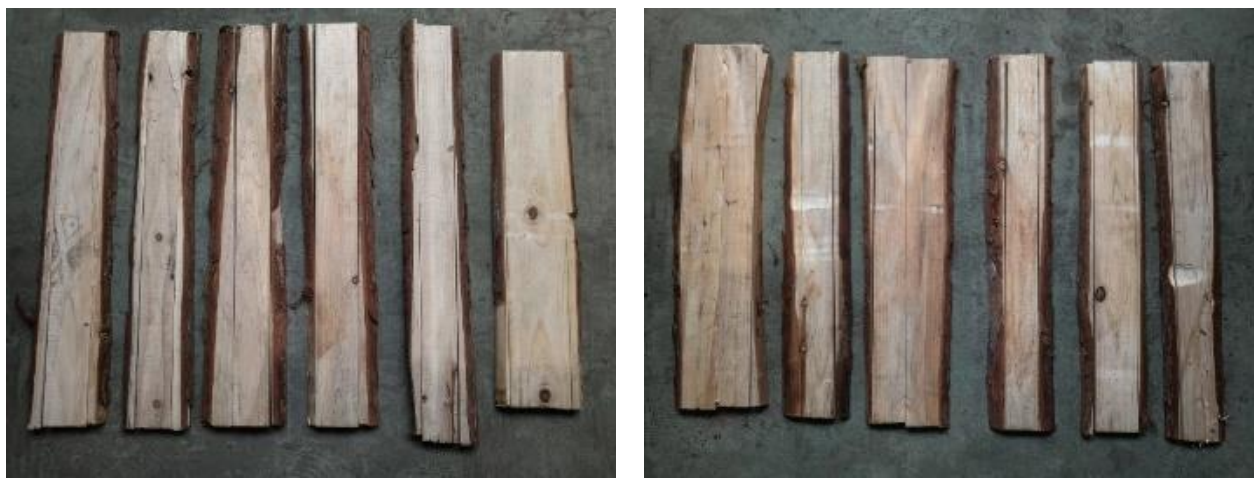

**Fig. 15. Examples of the bark removal results.**

## 6. Conclusion

To address the practical needs of sawmills regarding bark removal equipment, and to enhance the quality and efficiency of wood panels bark removal, this study explores and reveals the potential applications of deep learning in bark removal processes, leveraging its advantages in object detection and segmentation. A deep learning-based wood panels bark removal equipment was designed and developed, equipped with a custom-designed visual inspection system capable of capturing images of wood panels with bark in real-time. The captured images undergo accurate identification and segmentation of the wood panel areas using the BiSeNetV1 semantic segmentation model deployed on an industrial control computer. Subsequently, necessary key data for the bark removal process are calculated, enabling precise and efficient bark removal. Experimental research conducted thereafter fully validates the feasibility and rationale of the applied model, the accuracy of key data calculations, and the precision and efficiency of bark removal processes in actual sawmill production environments.

Throughout the development of the equipment, this study established a comprehensive semantic segmentation dataset for wood panels bark removal processing, which can provide data support for researchers engaged in wood processing-related studies and significantly promote the practical application of deep learning methods in this field. This work successfully realizes the application of deep learning technology in the timber industry, particularly in wood panels bark removal processing. It offers valuable references for subsequent application research by researchers, engineers, and related enterprises in this domain. However, despite the significant advantages demonstrated by the equipment developed in this study during technical validation and its practical application in bark removal processing at the sawmill, several challenges remain due to the diversity of wood panel characteristics and variations in production across different sawmills. For instance, the dataset established does not encompass all wood panel types, which presents difficulties when the model encounters wood panels with characteristics that differ significantly from those in the dataset. Future research will need to enhance the dataset's diversity by collecting images under various angles, lighting conditions, backgrounds, and environmental factors to improve the model's generalization capability. Additionally, the developed wood panel bark removal equipment is primarily designed for handling wood panels with bark ranging from 50 cm to 60 cm in length and 3 cm to 20 cm in width; thus, it cannot process wood panels exceeding these dimensions. Future studies will further expand the equipment's capabilities to meet the demand for processing longer and wider wood panels in sawmill operations.

## Conflict of interest statement

There is no conflict of interest.